\documentclass[lettersize,journal]{IEEEtran}
\usepackage{amsmath,amsfonts}
\usepackage{amsmath,amssymb,bm}
\usepackage{mathrsfs}
\usepackage{algorithmic}
\usepackage{algorithm}
\usepackage{array}
\usepackage[caption=false,font=normalsize,labelfont=sf,textfont=sf]{subfig}
\usepackage{textcomp}
\usepackage{stfloats}
\usepackage{url}
\usepackage{verbatim}
\usepackage{graphicx}
\usepackage{cite}
\usepackage{multirow} 
\usepackage{booktabs}
\begin{document}

\title{STQE: Spatial-Temporal Attribute Quality Enhancement for G-PCC Compressed Dynamic Point Clouds}

\author{Tian Guo, Hui Yuan,~\IEEEmembership{Senior Member,~IEEE,} Xiaolong Mao, Shiqi Jiang, Raouf Hamzaoui,~\IEEEmembership{Senior Member,~IEEE}, and Sam Kwong,~\IEEEmembership{Fellow,~IEEE}
        % <-this % stops a space
\thanks{This work was supported in part by the National Natural Science Foundation of China under Grants 62222110 and 62172259, the Taishan Scholar Project of Shandong Province (tsqn202103001), the Shandong Provincial Natural Science Foundation under Grant ZR2022ZD38, and the OPPO Research Fund. \textit{(Corresponding author: Hui Yuan)}} 
\thanks{Tian Guo and Hui Yuan are with the School of Control Science and Engineering, Shandong University, Ji'nan, 250061, China, and also with the Key Laboratory of Machine Intelligence and System Control, Ministry of Education, Ji'nan, 250061, China (e-mail: guotiansdu@mail.sdu.edu.cn; huiyuan@sdu.edu.cn).}
\thanks{Xiaolong Mao and Shiqi Jiang are with the School of Software, Shandong University, Ji'nan, 250100, China, and also with the School of Control Science and Engineering, Shandong University, Ji'nan, 250061, China. (e-mail: xiaolongmao@mail.sdu.edu.cn; shiqijiang@mail.sdu.edu.cn).}
\thanks{Raouf Hamzaoui is with the School of Engineering and Sustainable Development, De Montfort University, LE1 9BH Leicester, UK. (e-mail: rhamzaoui@dmu.ac.uk).}
\thanks{Sam Kwong is with the Department of Computing and Decision Science,
 Lingnan University, Hong Kong (e-mail: samkwong@ln.edu.hk).}
}

\markboth{Journal of \LaTeX\ Class Files,~Vol.~14, No.~8, August~2021}%
{Guo \MakeLowercase{\textit{et al.}}:STQE: Spatial-Temporal Attribute Quality Enhancement for G-PCC Compressed Dynamic Point Clouds}

% % column for its text to clear the IEEEpubid mark.

\maketitle

\begin{abstract}
Very few studies have addressed quality enhancement for compressed dynamic point clouds. In particular, the effective exploitation of spatial-temporal correlations between point cloud frames remains largely unexplored. Addressing this gap, we propose a spatial-temporal attribute quality enhancement (STQE) network that exploits both spatial and temporal correlations to improve the visual quality of G-PCC compressed dynamic point clouds. Our contributions include a recoloring-based motion compensation module that remaps reference attribute information to the current frame geometry to achieve precise inter-frame geometric alignment, a channel-aware temporal attention module that dynamically highlights relevant regions across bidirectional reference frames, a Gaussian-guided neighborhood feature aggregation module that efficiently captures spatial dependencies between geometry and color attributes, and a joint loss function based on the Pearson correlation coefficient, designed to alleviate over-smoothing effects typical of point-wise mean squared error optimization. When applied to the latest G-PCC test model, STQE achieved improvements of 0.855 dB, 0.682 dB, and 0.828 dB in delta PSNR (\(\Delta\)PSNR), with Bjøntegaard Delta rate (BD-rate) reductions of -25.2\%, -31.6\%, and -32.5\% for the Luma, Cb, and Cr components, respectively.
\end{abstract}


\section{Introduction}
\IEEEPARstart{W}{ith} the rapid development of 3D sensing technology, 3D point clouds are becoming increasingly popular for representing 3D scenes as sets of points with geometric coordinates and attribute information such as color, reflectance, and normal vectors \cite{refb1,refb2,refb3,refb4,refb5}. Point clouds play a vital role in many fields, such as autonomous driving, immersive communication, and virtual reality \cite{refc1, refb6,refb7,refb8}. A highly detailed point cloud usually contains millions or even billions of points for high resolution representation \cite{refb9,refb10,refb11,refb12}. However, this large data volume significantly challenges storage and transmission. Therefore, highly efficient point cloud compression is an urgent task. 

To standardize point cloud compression technology, the Moving Picture Experts Group (MPEG) launched a call for proposals for point cloud compression in 2017 \cite{refb13} and subsequently proposed two standards, i.e., video-based point cloud compression (V-PCC) \cite{refb14} and geometry-based point cloud compression (G-PCC) \cite{refb15}. V-PCC converts 3D point clouds into 2D video representations and uses advanced video coding standards such as H.265/HEVC \cite{refb16} or H.266/VVC \cite{refb17} for compression. In contrast, G-PCC directly processes 3D point clouds in 3D space, with its second edition named Enhanced G-PCC. With the growing popularity of immersive communication and augmented reality where denser point clouds are required, a dedicated branch of G-PCC, known as Solid G-PCC \cite{refb18}, has also been developed. However, lossy point cloud compression inevitably leads to distortions. To improve the coding efficiency further, quality enhancement is an efficient solution.

Quality enhancement for compressed videos has achieved great success, especially with deep neural networks \cite{ref1, ref2, ref3, ref4, ref5, ref8, ref6, ref7}. Similarly, enhancing the quality of compressed point clouds has recently emerged as an important research focus. \cite{refb4, refb11, refa1, refa2, refa4, refa5, refa6, refa7, refa9, refa10, refa11, refa12, refa13}. However, compared to videos, point cloud quality enhancement faces entirely new challenges, mainly due to the irregular distribution of points. Specifically, point clouds consist of unstructured and inherently sparse points, which makes it difficult to exploit spatial and temporal correlations effectively. At present, most compressed point cloud quality enhancement methods focus on single-frame static point clouds, lacking effective use of inter-frame correlations to achieve further gains. For dynamic point cloud sequences, variations in the number of points across frames and coordinate differences pose significant challenges for inter-frame motion compensation \cite{refe1}, which hinders the effective exploitation of temporal correlations.

We propose a spatial-temporal quality enhancement (STQE) method for G-PCC compressed dynamic point clouds by efficiently exploiting the spatial-temporal correlations between point cloud frames. STQE extracts temporal and spatial features through a bidirectional inter-frame feature extraction (BIFE) branch and a spatial feature extraction (SFE) branch, respectively, and then fuses the extracted features using a spatial-temporal feature fusion (STF) module. In the BIFE branch, we propose a simple yet effective recoloring-based motion compensation strategy that avoids explicit inter-frame motion estimation between point cloud frames, which is time consuming and operationally complex. This strategy accurately aligns inter-frame geometry, addressing challenges caused by varying numbers of points and complex inter-frame motion. In addition, to efficiently capture spatial features, we design a dense feature extraction block in the SFE branch based on a Gaussian-guided neighborhood feature aggregation (GNFA) module. In detail, the contributions of this paper are summarized as follows.
\begin{itemize}
\item{We propose an end-to-end spatial-temporal quality enhancement neural network for G-PCC compressed dynamic point clouds. This network consists of a BIFE branch, an SFE branch, and an STF module to efficiently extract and fuse spatial-temporal features.}
\item{We propose a recoloring-based motion compensation method in the BIFE branch that projects the color of a reference frame onto the geometry of the current frame to generate a virtual reference frame that is geometrically identical to the current frame. To further extract temporal features, we generate a forward and a backward virtual reference frame and design a channel-aware temporal attention module that dynamically focuses on the similarity between the current frame and these virtual reference frames.}
\item{We propose a GNFA module in the SFE branch, which uses a Gaussian kernel to adaptively weight the spatial neighborhood features of each point based on the statistical correlation between spatial distances and color attributes. This approach enhances the network's ability to capture spatial correlations.}
\item{We propose a joint loss function that uses the Pearson correlation coefficient as supplementary supervision to effectively restore high-frequency details.}
\end{itemize}

The remainder of this paper is organized as follows. Section II provides a brief review of related work. Section III describes the proposed method. Section IV presents and analyzes experimental results and analyses. Section V summarizes the main contributions and suggests future work. 

\section{Related Work}
Although the data structure of videos and point clouds are different, the quality enhancement methods for compressed videos can also inspire the design of quality enhancement for point clouds. Therefore, we review relevant work for both compressed video quality enhancement and compressed point cloud quality enhancement.
\subsection{Compressed video quality enhancement}
Yang et al. \cite{ref1} first designed a multi-frame quality enhancement (MFQE) method for the quality enhancement of HEVC compressed video. MFQE uses a support vector machine-based detector to identify peak quality frames (PQFs) and a multi-frame convolutional neural network to enhance the non-PQFs with the information of a pair of neighboring PQFs. Later, Guan et al. \cite{ref2} advanced MFQE and proposed MFQE2.0 by introducing the multi-scale strategy, batch normalization, and dense connection \cite{ref45}. Xiao et al. \cite{ref3} proposed a fast multi-scale deep decoder which uses a multi-scale 3D convolutional neural network (CNN) to explore multi-scale similarities between video frames to improve the quality of HEVC compressed videos. Meng et al. \cite{ref4} proposed a multi-frame guided attention network that integrates a motion flow module and temporal encoder to capture temporal variations and incorporates a partitioned average image for spatial guidance, which are then fused by a multi-scale guided encoder-decoder subnet to reconstruct high-quality video frames. Ding et al. \cite{ref5} proposed a patch-wise spatial-temporal quality enhancement network, which is capable of adaptively utilizing and enhancing compressed patches with both spatial and temporal information. More recently, they \cite{ref8} proposed a blind quality enhancement method for compressed videos, which exploits the fluctuated temporal information, feature similarity and feature difference between multiple quantization parameters (QPs) to achieve smooth quality among video frames. Moreover, Wang et al. \cite{ref6} proposed a generative adversarial network based on multi-level wavelet packet transform to exploit high-frequency details for enhancing the perceptual quality of compressed video. Luo et al. \cite{ref7} proposed a spatial-temporal detail information retrieval method for compressed video quality enhancement. They recovered temporal and spatial details using a multi-path deformable alignment module, several residual dense blocks, and channel attention mechanism. 

In summary, the performance of compressed video quality enhancement is mainly derived from two aspects: efficient spatial-temporal feature extraction achieved through multiscale analysis and spectrum analysis, and adaptive attention to regions with different levels of distortion achieved through attention mechanisms.

\subsection{Compressed point cloud quality enhancement}
Recent works for quality enhancement of compressed point cloud can be divided into two categories: quality enhancement of V-PCC and G-PCC compressed point clouds. 

For quality enhancement of V-PCC compressed point clouds, Akhtar et al. \cite{refa11} presented the first deep learning-based point cloud geometry compression artifact removal method. They used a projection-aware 3D sparse convolutional network to learn an embedding and then regresses over this embedding to learn the quantization noise. Xing et al. \cite{refa12} proposed a U-Net-based quality enhancement method for color attributes of dense 3D point clouds. In their approach, 3D patches are first generated from a distorted point cloud and then converted into 2D images using a specific scan order of points. These 2D images are subsequently enhanced using a U-Net-inspired neural network to improve their quality. Gao et al. \cite{refa13} proposed an occupancy-assisted compression artifact removal network to remove the distortion of attribute images at the decoder of V-PCC, which uses a multi-level feature fusion framework with channel-spatial attention based residual blocks to aggregate the occupancy information.
\begin{figure*}
\centering
\includegraphics[width=6.6in]{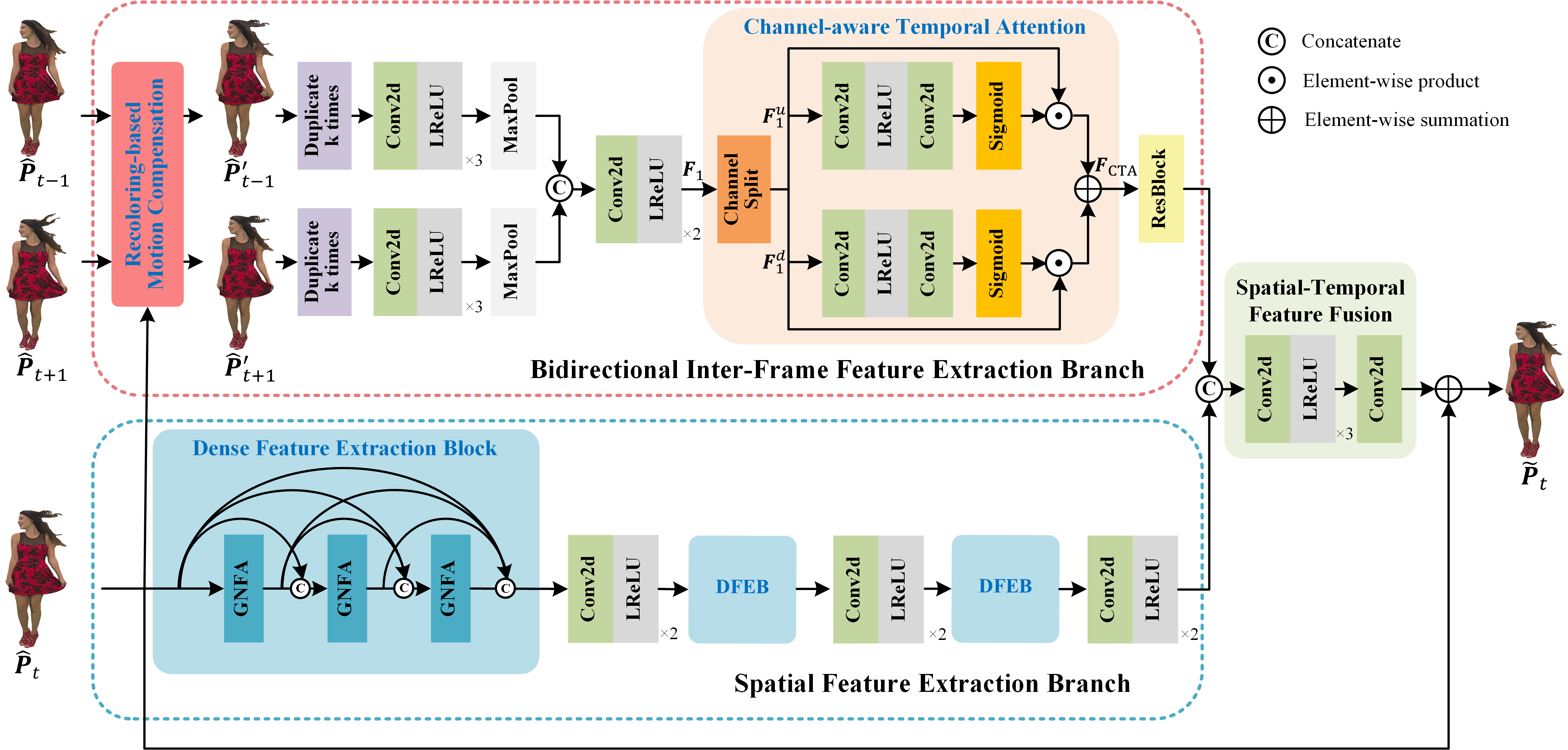}
\caption{Framework of the proposed STQE, where \({\widehat{\bm{P}}}_t\) denotes the \(t^{th}\) reconstructed frame, \({\widehat{\bm{P}}}_{t-1}\) and \({\widehat{\bm{P}}}_{t+1}\) denote the forward and backward reference frame of \({\widehat{\bm{P}}}_t\), respectively, and \({\widetilde{\bm{P}}}_t\) denotes the enhanced frame. DFEB: Dense Feature Extraction Block.}
\label{FIG1}
\end{figure*}

For quality enhancement of G-PCC compressed point clouds, Sheng et al. \cite{refa1} proposed a multi-scale graph attention network that constructs a geometry-assisted graph to treat point cloud attributes as graph signals and used Chebyshev graph convolutions to extract features and thus remove compression induced attribute artifacts. Xing et al. \cite{refa2} introduced a graph-based quality enhancement network that uses geometry information as an auxiliary input and graph convolution blocks to extract local features efficiently. Moreover, it can handle point clouds with various levels of distortion using a single pre-trained model. Zhang et al. \cite{refa4} proposed G-PCC++ that separately restores the geometry and attribute information of decoded point clouds by using the valid neighbors of each point in a local neighborhood. Later, they \cite{refa5} studied a fully data-driven approach and a rules-unrolling-based optimization for quality enhancement of G-PCC compressed point clouds. Moreover, they \cite{refa6} also designed a neural network consisting of two consecutive processing phases: multiple most probable sample offsets (MPSOs) derivation and MPSOs combination, for efficient quality enhancement. Tao et al. \cite{refa7} proposed a joint geometry and color hole repairing method, which uses a multi-view projection-based triangular hole detection scheme based on depth distribution to effectively repair the holes in both geometry and color of G-PCC compressed point clouds. Kathariya et al. \cite{refa9} extended the VVC Transformer-based spatial and frequency-decomposed feature fusion network (TSF-Net) into 3D domain for point clouds, proposing TSF-Net3D that uses sparse convolutions and channel-wise transformer-based multi-scale feature fusion to enhance the quality of color attribute. We \cite{refb11} proposed a Wiener filter-based method to effectively mitigate distortion accumulation during the coding process and enhance reconstruction quality.

However, the above methods are mainly applicable to single-frame static point clouds. For G-PCC dynamic point clouds, Wei et al. \cite{refb4} proposed a coefficients inheritance-based Wiener filter (CIWF), with Morton code-based fast nearest neighbor search, for inter-coded frames. Besides, Liu et al. \cite{refa10} proposed a deep learning-based quality enhancement method, namely DAE-MP, which uses an inter-frame motion prediction module to explicitly estimate motion displacement for inter-frame feature alignment. However, CIWF must embed the encoder to compute and transmit the filter coefficients, while DAE-MP only provides models for the Luma (Y) component at low bitrates and requires staged training, limiting their performance and applications. Therefore, we propose an end-to-end learning-based spatial-temporal attribute quality enhancement method for G-PCC compressed dynamic point cloud in this paper.

\section{Proposed Method}
Let \(\{\widehat{\bm{P}}_{t-N}, \ldots, \widehat{\bm{P}}_{t}, \ldots, \widehat{\bm{P}}_{t+N}\}\) be a reconstructed point cloud sequence, where \(\widehat{\bm{P}}_{t}=[\bm{P}^G_{t},\widehat{\bm{P}}^A_{t}]\) denotes the \(t^{th}\) frame with geometry \(\bm{P}^G_{t}\) and attribute \(\widehat{\bm{P}}^A_{t}\). The goal of attribute quality enhancement is to restore \(\{\widehat{\bm{P}}_{t-N}, \ldots, \widehat{\bm{P}}_{t}, \ldots, \widehat{\bm{P}}_{t+N} \}\) to an enhanced version \(\{ \widetilde{\bm{P}}_{t-N}, \ldots, \widetilde{\bm{P}}_{t}, \ldots, \widetilde{\bm{P}}_{t+N} \}\), where {\(\widetilde{\bm{P}}_{t}=[\bm{P}^G_{t},\widetilde{\bm{P}}^A_{t}]\), under the supervision of the original point cloud \(\{ {\bm{P}}_{t-N}, \ldots, {\bm{P}}_{t}, \ldots, {\bm{P}}_{t+N} \}\), where \({\bm{P}}_{t}=[\bm{P}^G_{t},{\bm{P}}^A_{t}]\). In the proposed STQE, we use the forward and backward frame of the current frame as reference frames, i.e., \(N = 1\). Therefore, the enhanced version \(\widetilde{\bm{P}}_{t}\) can be represented as

\begin{equation}
\label{1}
\widetilde{\bm{P}}_t = \Psi( \widehat{\bm{P}}_t, \widehat{\bm{P}}_{t-1}, \widehat{\bm{P}}_{t+1} \mid \mathbf{\Theta} ),
\end{equation}
where \(\Psi(\cdot)\) denotes the proposed STQE and \(\mathbf{\Theta}\) denotes the learnable parameters. 

To jointly use spatial and temporal information, the proposed STQE consists of BIFE branch (Section III-A) for temporal feature extraction, SFE branch (Section III-B) for spatial feature extraction, and STF module (Section III-C) for feature fusion, as shown in Fig. 1.

\subsection{Bidirectional Inter-frame Feature Extraction (BIFE)}
The inputs to the BIFE branch include the current frame \(\widehat{\bm{P}}_t\), its forward and backward reference frames \(\widehat{\bm{P}}_{t-1}\) and \(\widehat{\bm{P}}_{t+1}\), to effectively extract the temporal correlation between adjacent frames. First, \(\widehat{\bm{P}}_t\), \(\widehat{\bm{P}}_{t-1}\), and \(\widehat{\bm{P}}_{t+1}\) are fed into the recoloring-based motion compensation (RMC) module to generate virtual reference frames \(\widehat{\bm{P}}^\prime_{t-1}\) and \(\widehat{\bm{P}}^\prime_{t+1}\) to align inter-frame geometry, thus laying the foundation for the subsequent extraction of temporal domain features. Then, \(\widehat{\bm{P}}^\prime_{t-1}\) and \(\widehat{\bm{P}}^\prime_{t+1}\) are respectively duplicated \(k\) times and input to three 2D convolutional layers with a kernel size of \(1\times1\), and the Leaky ReLU activation function, to extract shallow local spatial features. In addition, the max pooling operation is applied to further filter key features. The obtained features are concatenated and then input into two convolutional layers to integrate complementary information from forward and backward frames to obtain the feature \(\bm{F}_1\). Next, \(\bm{F}_1\) is fed into the channel-aware temporal attention (CTA) module to adaptively select reference regions with stronger correlation with \(\widehat{\bm{P}}_t\) on different channels, to efficiently use reference information and obtain feature \(\bm{F}_{CTA}\). Finally, \(\bm{F}_{CTA}\) is further refined by the ResBlock to obtain the final temporal feature. The details of RMC module, CTA module, and ResBlock are as follows.

\noindent\textit{\textbf{1) Recoloring-based Motion Compensation (RMC)}}

The change in the number of points and the coordinate difference between frames in the dynamic point cloud sequence make motion estimation and compensation difficult to operate. However, the RMC module avoids explicit motion estimation and directly remaps the reference frame color to the geometry coordinates of the current frame, achieving complete alignment of the geometry coordinates between frames and eliminating the color misalignment problem caused by inter-frame motion. As shown in Fig. 2, the current frame \(\widehat{\bm{P}}_{t}=[\bm{P}^G_{t},\widehat{\bm{P}}^A_{t}]\) and the backward reference frame \(\widehat{\bm{P}}_{t+1}=[\bm{P}^G_{t+1},\widehat{\bm{P}}^A_{t+1}]\) are taken as an example to show the complete process of RMC module. First, the geometry of the virtual reference frame is specified by \(\bm{P}^G_{t}\). Second, each point in \(\bm{P}^G_{t+1}\) is traversed to find its nearest neighbor point in \(\bm{P}^G_{t}\) by the \(k\)-nearest neighbor (KNN) search algorithm \cite{refb19}, and directly mapped onto the nearest neighbor point in the virtual reference frame. When the current and reference frames contain different numbers of points, each point is processed as follows: if one or more reference points are mapped to it, their average attribute is assigned to that point; otherwise, the current point retains its original attribute. As a result, the virtual reference frame \(\widehat{\bm{P}}^\prime_{t+1}=[\bm{P}^G_{t},\widehat{\bm{P}}^{{\prime}A}_{t+1}]\) can be obtained. The same operation is performed for the forward reference frame \(\widehat{\bm{P}}_{t-1}=[\bm{P}^G_{t-1},\widehat{\bm{P}}^A_{t-1}]\) as well to obtain \(\widehat{\bm{P}}^\prime_{t-1}=[\bm{P}^G_{t},\widehat{\bm{P}}^{{\prime}A}_{t-1}]\). 

\noindent\textit{\textbf{2) Channel-aware Temporal Attention (CTA)}}

Different from equally using of adjacent frames, CTA module accurately extracts more valid reference information by adaptively giving higher weights to the reference regions that are more relevant to the current frame in particular local regions or channels. Moreover, CTA enhances the model’s representational capacity through channel splitting, enabling it to compute attention weights independently for each reference frame. Specifically, as shown in Fig. 1, for feature \(\bm{F}_1 \in \mathbb{R}^{n \times c}\), where \(n\) represents the number of points and \(c\) is the feature dimension, we first split it into two independent parts, \(\bm{F}^u_1 \in \mathbb{R}^{n \times (c/2)}\) and \(\bm{F}^d_1 \in \mathbb{R}^{n \times (c/2)}\), corresponding to the forward and backward reference frames, and then combine two convolution layers with a kernel size of \(1\times1\) and a Leaky ReLU activation function to obtain the temporal-wise dependencies, and finally dynamically generate temporal-wise attention weights through sigmoid activation function, and obtain \(\bm{F}_{CTA} \in \mathbb{R}^{n \times (c/2)}\) by
\begin{equation}
\label{2}
\bm{F}_{{CTA}} = \bm{F}_1^{u} \odot \mathcal{\bm{S}} ( \bm{W}_{2d} ( \bm{F}_1^{u} ) ) + \bm{F}_1^{d} \odot \mathcal{\bm{S}} ( \bm{W}_{2d} ( \bm{F}_1^{d} ) ),
\end{equation}
where \(\mathcal{\bm{S}}(\cdot)\) denotes the Sigmoid function, \(\bm{W}_{2d}(\cdot)\) denotes the combination of two convolution layers with a kernel size of \(1\times1\) and a Leaky ReLU activation function, and \(\odot\) denotes element-wise product.
\begin{figure}
\centering
\includegraphics[width=3.3in]{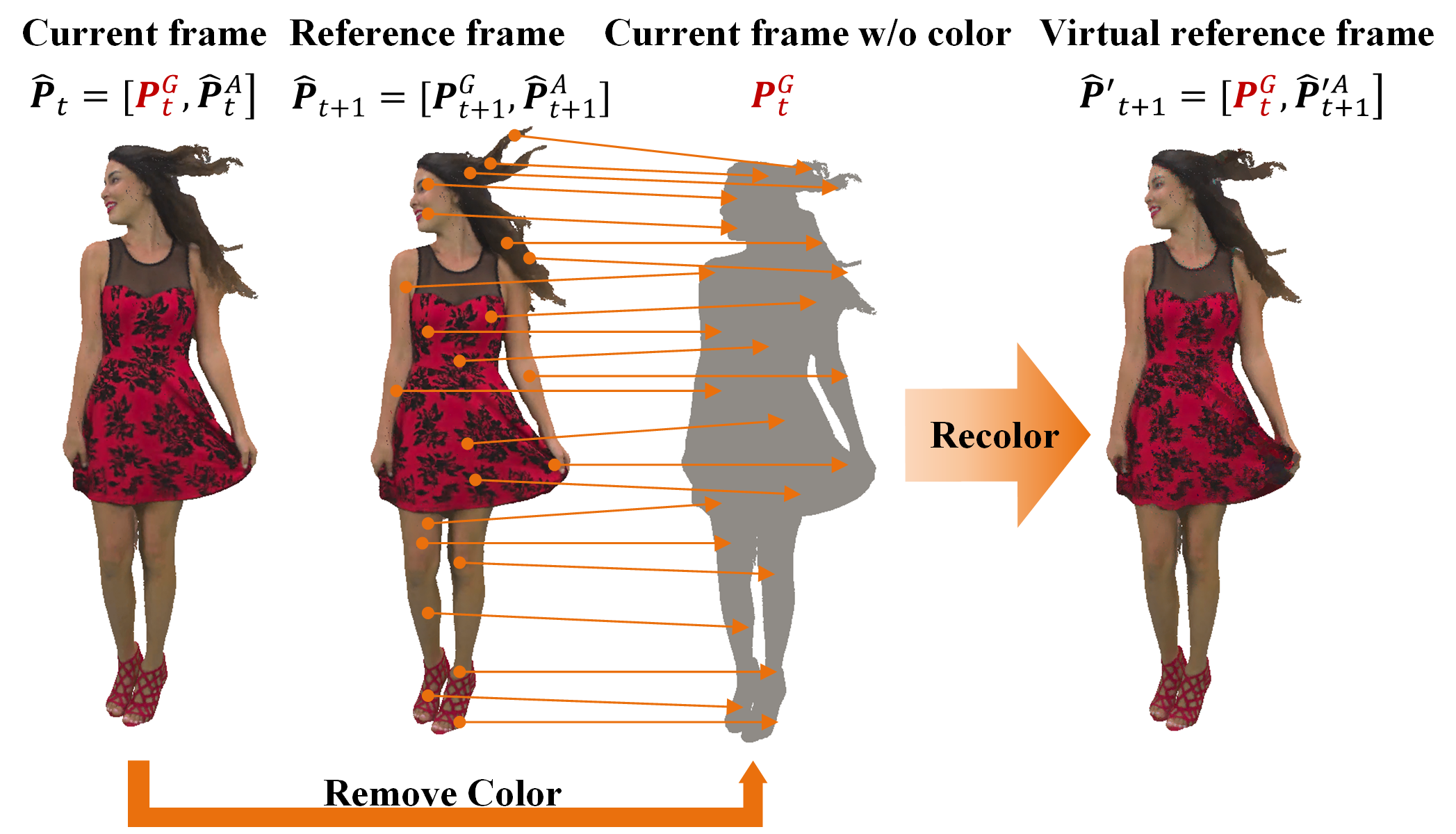}
\caption{Framework of the RMC module.}
\label{FIG2}
\end{figure}

\noindent\textit{\textbf{3) ResBlock}}

ResBlock consists of four \(1\times1\) convolution layers interleaved with three Leaky ReLU activation functions and strengthened by residual connections, aiming to further extract deep temporal features.

\begin{figure*}
\centering
\includegraphics[width=6.5in]{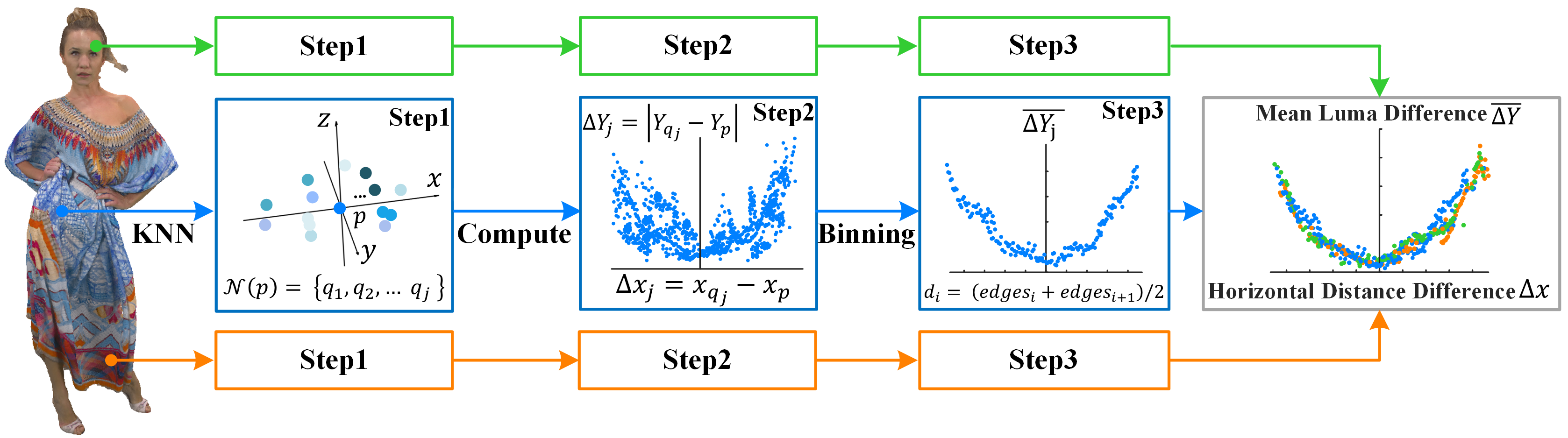}
\caption{Flowchart of the experiment for illustrating spatial correlation among points. Colored arrows indicate three repetitions of the experiment (Steps 1–3). Each color denotes one repetition.}
\label{FIG3}
\end{figure*}

\begin{figure}
\centering
\includegraphics[width=3.4in]{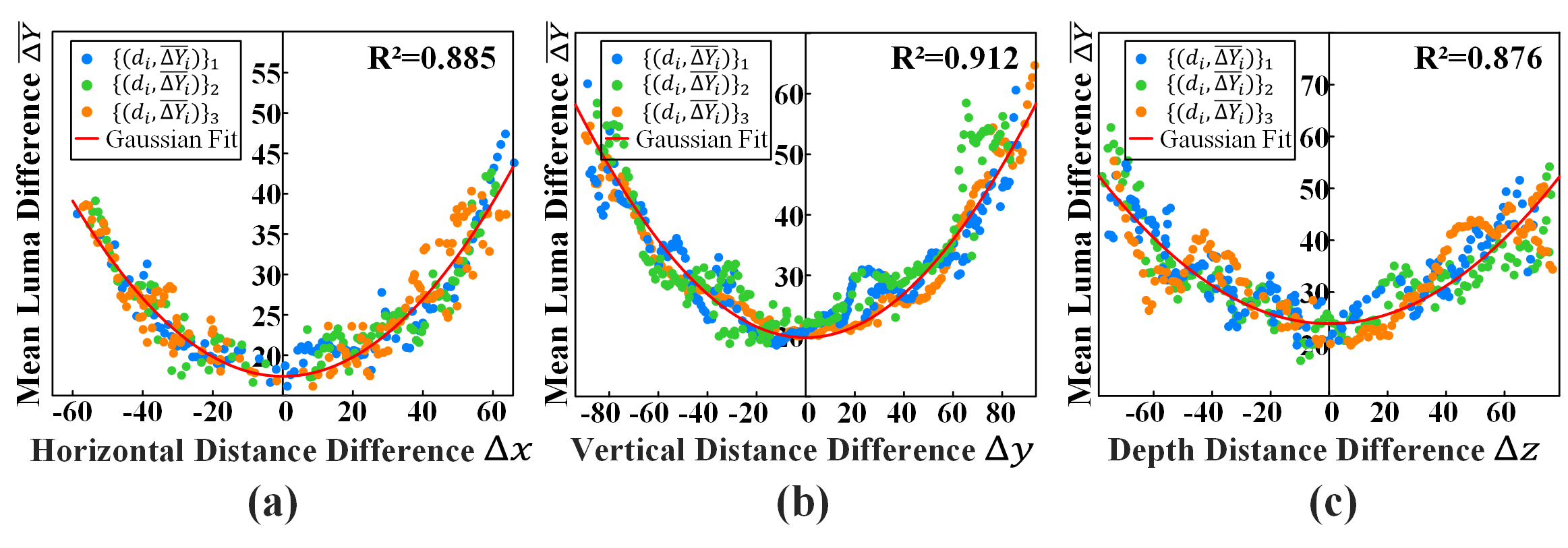}
\caption{Fitted Gaussian model and the actual plot between the horizontal distance difference \(\Delta x\) and the mean of the luma component \(\overline{\Delta Y}\), the vertical distance difference \(\Delta y\) and \(\overline{\Delta Y}\), and the depth distance difference \(\Delta z\) and \(\overline{\Delta Y}\), of the point cloud \textit{longdress\_vox10}.}
\label{FIG4}
\end{figure}

\begin{figure*}
\centering
\includegraphics[width=6.5in]{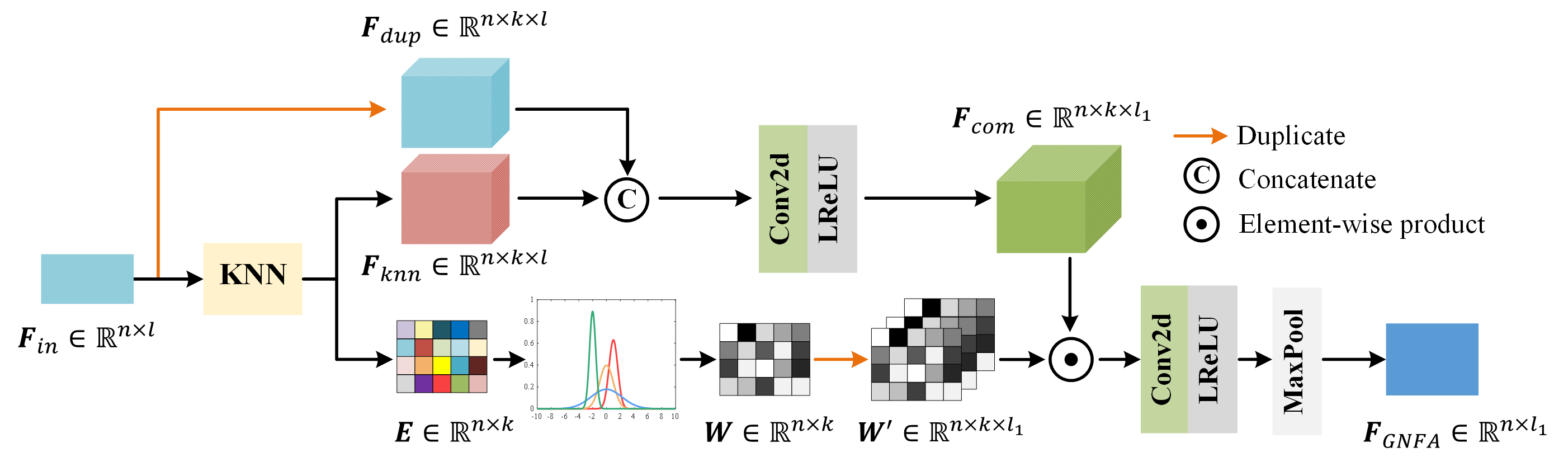}
\caption{Framework of the GNFA module.}
\label{FIG5}
\end{figure*}
\subsection{Spatial Feature Extraction (SFE)}
To extract spatial features, we propose a dense feature extraction block, consisting of three densely connected GNFA modules as described below.

\noindent\textit{\textbf{1) Gaussian-guided Neighborhood Feature Aggregation (GNFA)}}

To statistically illustrate the spatial correlation among points, we conducted the following experiments as shown in Fig. 3.

Step 1: Taking the point cloud \textit{longdress\_vox10} as an example, we randomly select one point \(p\), whose coordinates and luma component are \((x_p,y_p,z_p)\) and \(Y_p\), and use the KNN algorithm to find the \(g\) nearest neighbours \(q_j\) of this point to obtain the set of nearest neighbors \(\mathcal{N}(p) = \{ q_1, q_2, \ldots, q_j \}, j = 1,2,\ldots,g\). 

Step 2: For each nearest neighbour \(q_j\), we compute
\begin{equation}
\label{3}
\left\{
\begin{aligned}
\Delta x_j &= x_{q_j} - x_p \\
\Delta Y_j &= \left| Y_{q_j} - Y_p \right|,
\end{aligned}
\right.
\end{equation}
where \(\Delta x_j\) is the horizontal distance and \(\Delta Y_j\) is the absolute difference between \(Y_{q_j}\) and \(Y_p\). 

Step 3: We find the minimum (\(d_{min}\)) and maximum (\(d_{max}\)) of the horizontal distances,
\begin{equation}
\label{4}
\left\{
\begin{aligned}
d_{min} &= \min\limits_{j} \Delta x_j \\
d_{max} &= \max\limits_{j} \Delta x_j,
\end{aligned}
\right.
\end{equation}
and partition the interval \([d_{min},d_{max}]\) into uniform bins of width \(\Delta d=0.5\) to obtain the bin edges, 
\begin{equation}
\label{5}
\textit{edges} = [ d_{min},\ d_{min} + \Delta d,\ d_{min} + 2\Delta d,\ \ldots,\ d_{max} ].
\end{equation}
Next, for the \(i^{th}\) bin, we define its center as
\begin{equation}
\label{6}
d_i = \frac{\textit{edges}_i + \textit{edges}_{i+1}}{2},
\end{equation}
and compute the mean \((\overline{\Delta Y_j})\) of all \(\Delta Y_j\) falling in that bin to get the pair data \((d_i,\overline{\Delta Y_j})\).

Finally, we repeat the above steps for \(N_{runs}=3\) times to obtain three sets of binned pair data \((d_i,\overline{\Delta Y_j})_{N_{runs}}\) represented by green, blue, and orange points, and plot them with the same coordinate system. All the pair data points are pooled and fitted by a Gaussian model via nonlinear least squares, as shown in Fig. 4(a) where the squared correlation coefficient \(R^2\) between the raw data and the fitted data is also shown. We can see that \(R^2=0.885\), indicating an accurate Gaussian decay relationship between the horizontal distances and the mean difference of luma components in the point cloud. Similarly, the process is repeated in the vertical and depth directions, respectively. As shown in Fig. 4 (b) and (c), the corresponding \(R^2\) are 0.912 and 0.876, respectively, which further verifies the accuracy of the Gaussian decay trend between the distance and luma difference of points in different directions. Based on this statistical conclusion, we then propose the GNFA module as shown in Fig. 5.

For the input feature \(\bm{F}_{in} \in \mathbb{R}^{n \times l}\), where \(n\) denotes the number of points and \(l\) is the feature dimension, first, it is duplicated \(k\) times to obtain the feature \(\bm{F}_{dup} \in \mathbb{R}^{n \times k \times l}\). Simultaneously, the KNN algorithm is used to search for the \(k\) nearest neighbors of each point to obtain the feature \(\bm{F}_{knn} \in \mathbb{R}^{n \times k \times l}\) and the corresponding Euclidean distance matrix \(\bm{E} \in \mathbb{R}^{n \times k}\) whose element is \(e_{ij}\), \(i\in[1,n]\), \(j\in[1,k]\). Second, \(\bm{F}_{dup}\) and \(\bm{F}_{knn}\) are concatenated together, and a 2D convolution with \(1\times1\) kernel and LeakyReLU is applied to obtain the feature \(\bm{F}_{com} \in \mathbb{R}^{n \times k \times l_1}\) that embeds neighbor points’ information into each point. Subsequently, a neighborhood weight matrix \(\bm{W} \in \mathbb{R}^{n \times k}\) whose element is \(w_{ij}\) is defined based on the Gaussian kernel
\begin{equation}
\label{7}
w_{ij} = \exp\left( -\frac{e_{ij}}{2\sigma^2} \right),
\end{equation}
where \(\sigma^2\) denotes a kernel bandwidth parameter controlling the decay rate, which is empirically set to 0.5. \(w_{ij}\) tends to 1 as \(e_{ij}\) tends to 0, and \(w_{ij}\) tends to 0 as \(e_{ij}\) tends to its maximum value. Afterwards, \(\bm{W}\) is duplicated \(l_1\) times to get \(\bm{W}^\prime \in \mathbb{R}^{n \times k \times l_1}\), which is element-wise multiplied by \(\bm{F}_{com}\). Finally, the weighted features are fed into a 2D convolution with a \(1\times1\) kernel, LeakyReLU, and a max pooling layer, to obtain the feature \(\bm{F}_{GNFA} \in \mathbb{R}^{n \times l_1}\). 

Unlike traditional uniform-weighted neighborhood aggregation methods, GNFA exploits the relationship between inter-point distance and attribute differences to adaptively assign larger weights to the features of the neighborhood that have higher correlation with the current point, thus improving the feature expression ability of the network.
\subsection{Spatial-Temporal Feature Fusion (STF)}
The STF module fuses temporal and spatial features through a series of convolutional layers to capture joint information in the spatial-temporal domain. Specifically, STF takes temporal and spatial features as inputs, which are processed through a sequential structure consisting of three consecutive 2D convolutional layers with a \(1\times1\) kernel and LeakyReLU. The structure effectively strengthens the nonlinear mapping relationship between spatial-temporal features, extracts the deep fusion features. Finally, after a 2D convolution, the fused features are squeezed to the dimension of \(n\times1\), outputting the final distortion-aware features.
\subsection{Loss Function}
Existing methods usually use a point-wise mean loss, such as mean square error (MSE), to minimize the differences between the enhanced point cloud \({\widetilde{\bm{P}}}_t^A\) and the original point cloud \(\bm{P}_t^A\),

\begin{equation}
\label{8}
L_{{MSE}} = \frac{1}{n} \left\| \widetilde{\bm{P}}_t^{A} - \bm{P}_t^{A} \right\|_2^2,
\end{equation}
where \(n\) denotes the number of points. However, such a loss function often leads to excessive smoothing, potentially resulting in the loss of high-frequency details. To address this problem, we introduce a complementary loss that uses the Pearson correlation coefficient (PCC) to assess the loss of high-frequency spatial details. We denote this loss by \(L_{PCC}\) and compute it as:

\begin{equation}
\label{9}
L_{{PCC}} = 1 - \frac{\mathrm{Cov}( \widetilde{\bm{P}}_t^{A},\ \bm{P}_t^{A} )}{\sqrt{\mathrm{Var}( \widetilde{\bm{P}}_t^{A}) \cdot \mathrm{Var}( \bm{P}_t^{A} )}},
\end{equation}
where \(\mathrm{Cov}(\cdot)\) denotes covariance and \(\mathrm{Var}(\cdot)\) denotes variance. The proposed joint loss function is

\begin{equation}
\label{10}
L= L_{MSE} + \alpha L_{PCC},
\end{equation}
where \(\alpha\) is a trade-off hyper-parameter.

\section{Experimental Results and Analysis}
In Section IV-A, we introduce the experimental settings, including the dataset, implementation details, and evaluation metrics. In Section IV-B, we assess the enhanced point clouds in terms of objective quality and compare the coding efficiency before and after integrating the proposed method into G-PCC. In Section IV-C, we illustrate the robustness of the proposed STQE. In Section IV-D, we compare STQE with the state-of-the-art deep learning-based point cloud quality enhancement method. In Section IV-E, we conduct ablation studies to evaluate how each component of STQE contributes to the overall performance. In Section IV-F, we analyze the complexity of STQE.
\subsection{Experimental Setup}
\noindent\textit{\textbf{1) Datasets}}

We trained the proposed model using five dynamic point cloud sequences \textit{Longdress}, \textit{Basketball\_player}, \textit{Exercise}, \textit{Andrew}, and \textit{David}. \textit{Longdress} was taken from the 8i Voxelized Full Bodies dataset (8iVFB v2) \cite{refb20} with 10-bit precision. \textit{Basketball\_player} and \textit{Exercise} were taken from the Owlii Dynamic Human Textured Mesh Sequence dataset (Owlii) \cite{refb21} with 11-bit precision. \textit{Andrew} and \textit{David} were taken from the Microsoft Voxelized Upper Bodies dataset (MVUB) \cite{refb22} with 10-bit precision. The frame rate of each sequence is 30 fps. We encoded the sequences using the latest G-PCC Test Model Category 13 version 28.0 (TMC13v28) \cite{refb23}, applying inter-frame prediction mode with octree-RAHT configuration to generate training datasets. The encoding was conducted under the Common Test Condition (CTC) C1 \cite{refb24}, which involves lossless geometry compression and lossy attribute compression. We collected the first 32 frames of each sequence for training, a total of 160 frames. Due to limitations in GPU memory capacity, we used a patch generation-and-fusion approach in the same way as \cite{refa2}. 

We tested the performance of STQE on nine sequences: \textit{Loot, Redandblack, Soldier, Dancer, Model, Phil, Ricardo, Sarah,} and \textit{Queen}.\textit{ Loot, Redandblack}, and \textit{Soldier} were taken from the 8iVFB v2 dataset with 10-bit precision. Sequences \textit{Dancer} and \textit{Model} were taken from the Owlii dataset with 11-bit precision. \textit{Phil, Ricardo}, and \textit{Sarah} were taken from the MVUB dataset with 10-bit precision. \textit{Queen} was taken from the Technicolor dataset \cite{refb25} with 10-bit precision. The framerate of Queen is 50 fps, while that of all other sequences is 30 fps. Each sequence was compressed using TMC13v28 with quantization parameters (QPs) 51, 46, 40, 34, 28, 22, corresponding to the six bitrates, R01, R02, R03, R04, R05, and R06. We collected the first 32 frames of each sequence for testing, a total of 288 frames.

\begin{figure*}
\centering
\includegraphics[width=6.2in]{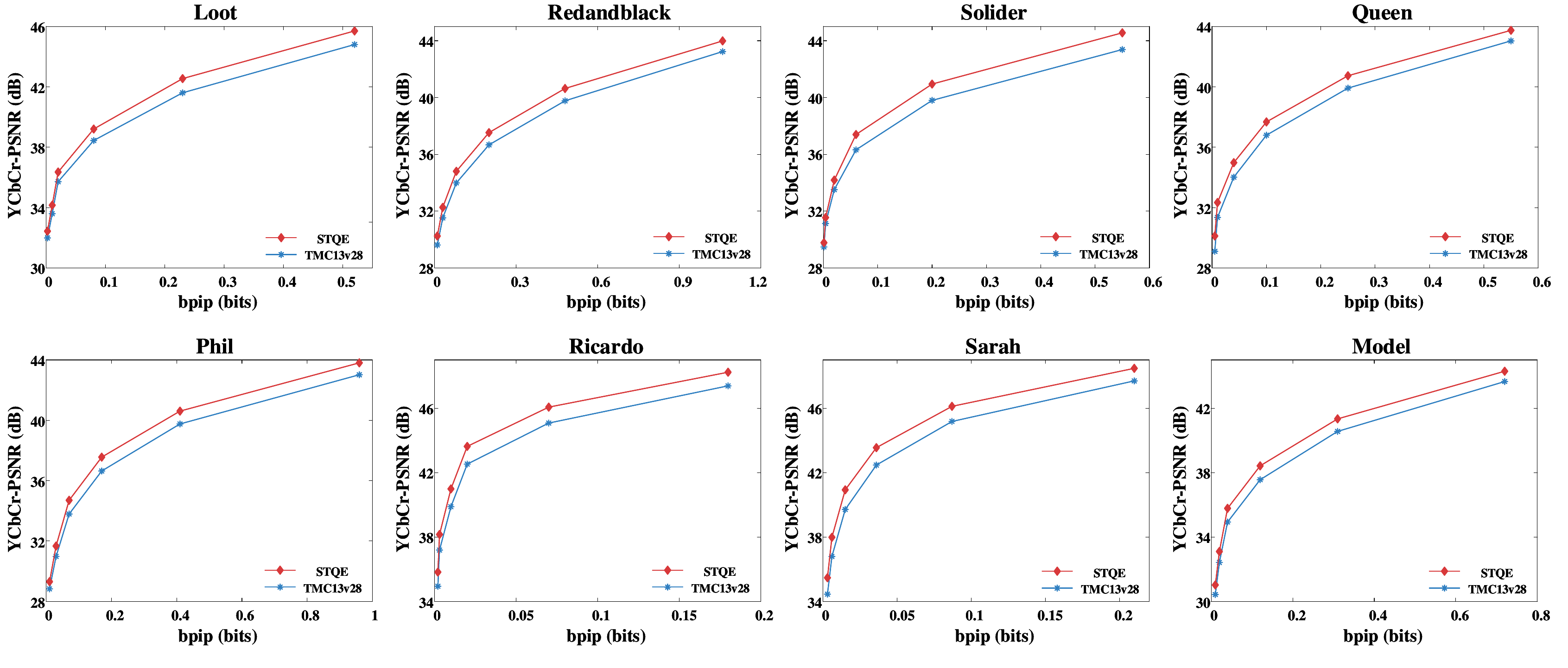}
\caption{Rate-PSNR curves before and after integrating STQE into G-PCC.}
\label{FIG6}
\end{figure*}
\begin{figure*}
\centering
\includegraphics[width=6.2in]{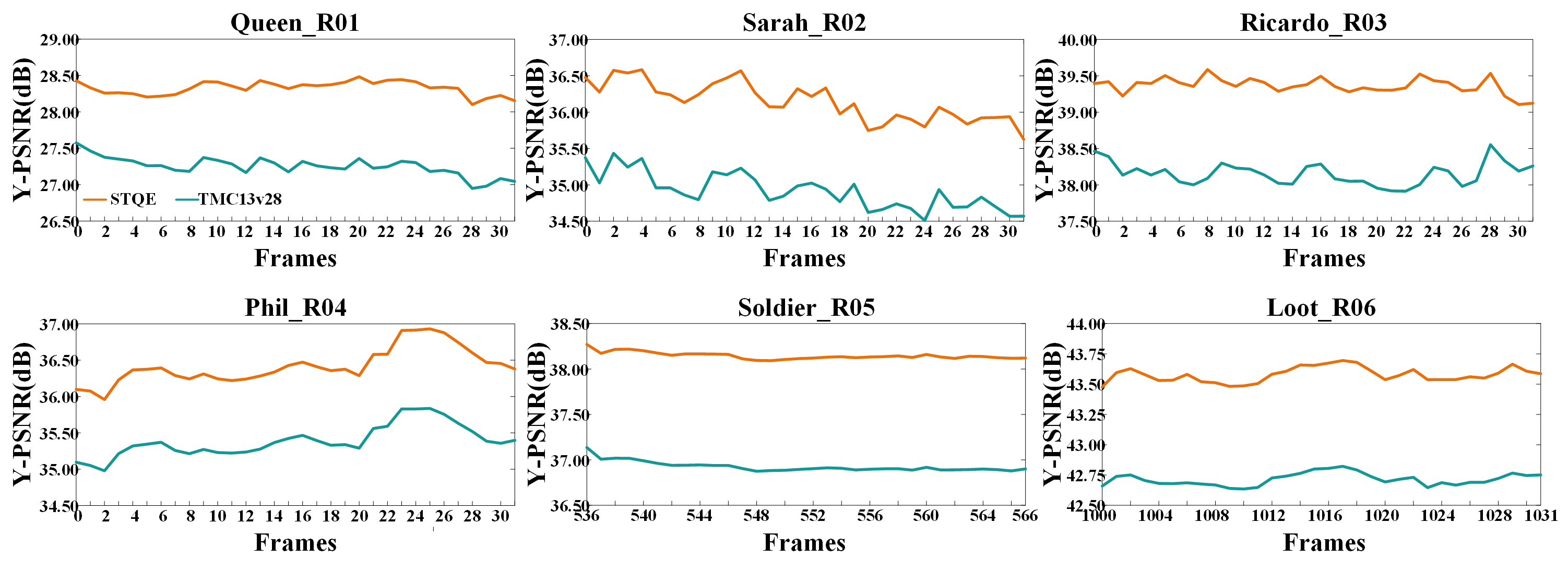}
\caption{PSNR curves of six test sequences at six bitrates before and after using STQE.}
\label{FIG7}
\end{figure*}

\begin{table}[ht]
\centering
\caption{\(\Delta\)PSNR (dB) AND BD-rate (\%) AFTER INTEGRATING STQE INTO G-PCC}
\label{tab:my_label1}  
\resizebox{8.55cm}{!}{
\begin{tabular}{lcccccccc}
\toprule
\multirow{2}{*}{\textbf{Sequence}} & \multicolumn{4}{c}{\textbf{\(\Delta\)PSNR (dB)}} & \multicolumn{4}{c}{\textbf{BD-rate (\%)}} \\ 
 & \textbf{Y} & \textbf{Cb} & \textbf{Cr} & \textbf{YCbCr} & \textbf{Y} & \textbf{Cb} & \textbf{Cr} & \textbf{YCbCr} \\ \hline
         Loot & 0.618 & 0.917 & 1.034 & 0.707 & -21.6 & -42.4 & -44.4 & -27.1 \\ 
        Redandblack & 0.786 & 0.678 & 0.829 & 0.778 & -23.7 & -30.3 & -20.3 & -24.1 \\ 
        Soldier & 0.850 & 0.614 & 0.675 & 0.798 & -27.6 & -34.1 & -34.7 & -29.3 \\ 
        Queen & 0.865 & 1.006 & 0.973 & 0.896 & -24.7 & -34.8 & -31.5 & -26.8 \\ 
        Dancer & 0.805 & 0.419 & 0.869 & 0.765 & -23.2 & -25.0 & -35.1 & -24.9 \\ 
        Model & 0.753 & 0.427 & 0.917 & 0.733 & -22.4 & -24.4 & -35.0 & -24.2 \\ 
        Phil & 0.881 & 0.387 & 0.502 & 0.772 & -23.3 & -19.0 & -20.8 & -22.5 \\ 
        Ricardo & 1.041 & 0.825 & 0.815 & 0.986 & -31.0 & -40.7 & -37.4 & -33.0 \\ 
        Sarah & 1.096 & 0.867 & 0.837 & 1.035 & -29.4 & -33.9 & -32.8 & -30.4 \\ \hline
        Average & \textbf{0.855} & \textbf{0.682} & \textbf{0.828} & \textbf{0.830} & \textbf{-25.2} & \textbf{-31.6} & \textbf{-32.5} & \textbf{-26.9} \\ \bottomrule
\end{tabular}}
\end{table}
\begin{table}
\centering
\caption{\(\Delta\)PSNR (dB) ACHIEVED BY STQE IN THE Y COMPONENT}
\label{tab:my_label3}  
\resizebox{8cm}{!}{
\begin{tabular}{lcccccc}
\toprule
\multirow{2}{*}{\textbf{Sequence}} & \multicolumn{6}{c}{\textbf{\(\Delta\)PSNR (dB)}} \\ 
 & \textbf{R01} & \textbf{R02} & \textbf{R03} & \textbf{R04} & \textbf{R05} & \textbf{R06}  \\ \hline
        Loot & 0.414 & 0.445 & 0.474 & 0.649 & 0.863 & 0.863 \\ 
        Redandblack & 0.646 & 0.806 & 0.840 & 0.830 & 0.845 & 0.746 \\ 
        Soldier & 0.315 & 0.469 & 0.715 & 1.202 & 1.220 & 1.177 \\ 
        Queen & 1.074 & 0.952 & 0.916 & 0.796 & 0.763 & 0.687 \\ 
        Dancer & 0.774 & 0.873 & 0.983 & 0.955 & 0.749 & 0.497 \\ 
        Model & 0.661 & 0.753 & 0.922 & 0.881 & 0.750 & 0.552 \\ 
        Phil & 0.539 & 0.806 & 1.046 & 1.033 & 0.978 & 0.886 \\ 
        Ricardo & 0.954 & 1.035 & 1.212 & 1.170 & 1.025 & 0.850 \\ 
        Sarah & 1.083 & 1.233 & 1.276 & 1.146 & 1.012 & 0.828 \\ \hline
        \textbf{Average} & \textbf{0.718} & \textbf{0.819} & \textbf{0.932} & \textbf{0.962} & \textbf{0.912} & \textbf{0.787} \\ \bottomrule
\end{tabular}}
\end{table}
\noindent\textit{\textbf{2) Implementation Details}}

We trained the proposed STQE for 50 epochs with a batch size of 16, an Adam optimizer \cite{refb26} with a learning rate of 0.0001. Moreover, we set \(k = 20\) in KNN algorithm and \(\alpha = 1\) in the loss function. We implemented the proposed method on an NVIDIA GeForce RTX4090 GPU, using PyTorch v1.12. We trained three models, corresponding to three color components (Y, Cb, and Cr). Each component was processed independently.

\noindent\textit{\textbf{3) Evaluation Metrics}}

We used the \(\Delta\)PSNR and BD-rate metrics \cite{refb27} to evaluate the performance of STQE compared to TMC13v28. \(\Delta\)PSNR measures the PSNR difference between the proposed method and the anchor at a single bitrate while the BD-rate measures the average bitrate increment in bits per input point (bpip) at the same PSNR when integrating the proposed STQE method into G-PCC encoder. A positive \(\Delta\)PSNR and a negative BD-rate indicate that the proposed method improved TMC13v28. In addition to calculating the PSNR for all the color components, we also used a combined PSNR, i.e., YCbCr-PSNR \cite{refq1}, which calculates the weighted average PSNR of Y, Cb, and Cr by a ratio of 6:1:1, to evaluate the overall color quality gains brought by the proposed method. 

\subsection{Objective Quality Evaluation}

Table I shows the \(\Delta\)PSNR and BD-rates achieved by STQE, averaged over the first 32 frames of each test sequence. We can see that STQE achieved average \(\Delta\)PSNR of 0.855 dB, 0.682 dB, 0.828 dB, and 0.830 dB for the Y, Cb, Cr components and combined YCbCr, respectively, corresponding to -25.2\%, -31.6\%, -32.5\%, and -26.9\% BD-rates, respectively. The largest PSNR gains were notably high, reaching 1.276 dB for the Y component of sequence \textit{sarah} at R03, 1.192 dB for the Cb component of sequence \textit{queen} at R03, and 1.241 dB for the Cr component of sequence \textit{redandblack} at R05. Table II shows the \(\Delta\)PSNRs of the Y component at six bitrates achieved by STQE. The gains were significant at all bitrates. The medium bitrates, R03 and R04, showed the highest improvements, where the average PSNR gains reached 0.932 dB and 0.962 dB, respectively. Fig. 6 compares the rate-PSNR curves before and after integrating STQE into G-PCC. The results show that the proposed method significantly improved the coding efficiency of G-PCC.

In addition, as shown in Fig. 7, we provided PSNR variations along with frame indexes of six test sequences at six bitrates before and after performing STQE. We can see that STQE can achieve significant improvements over all compressed frames.

% \subsection{Subjective Quality Evaluation}
% Taking sequences \textit{redandblack} and \textit{soldier} as an example, Fig. 8 compares the original point clouds in the first row, the point clouds compressed and reconstructed by G-PCC in the second row, and the point clouds enhanced by STQE in the third row. Applying STQE significantly enhanced subjective quality, notably improving texture clarity and color transitions.

\begin{figure*}
\centering
\includegraphics[width=6.2in]{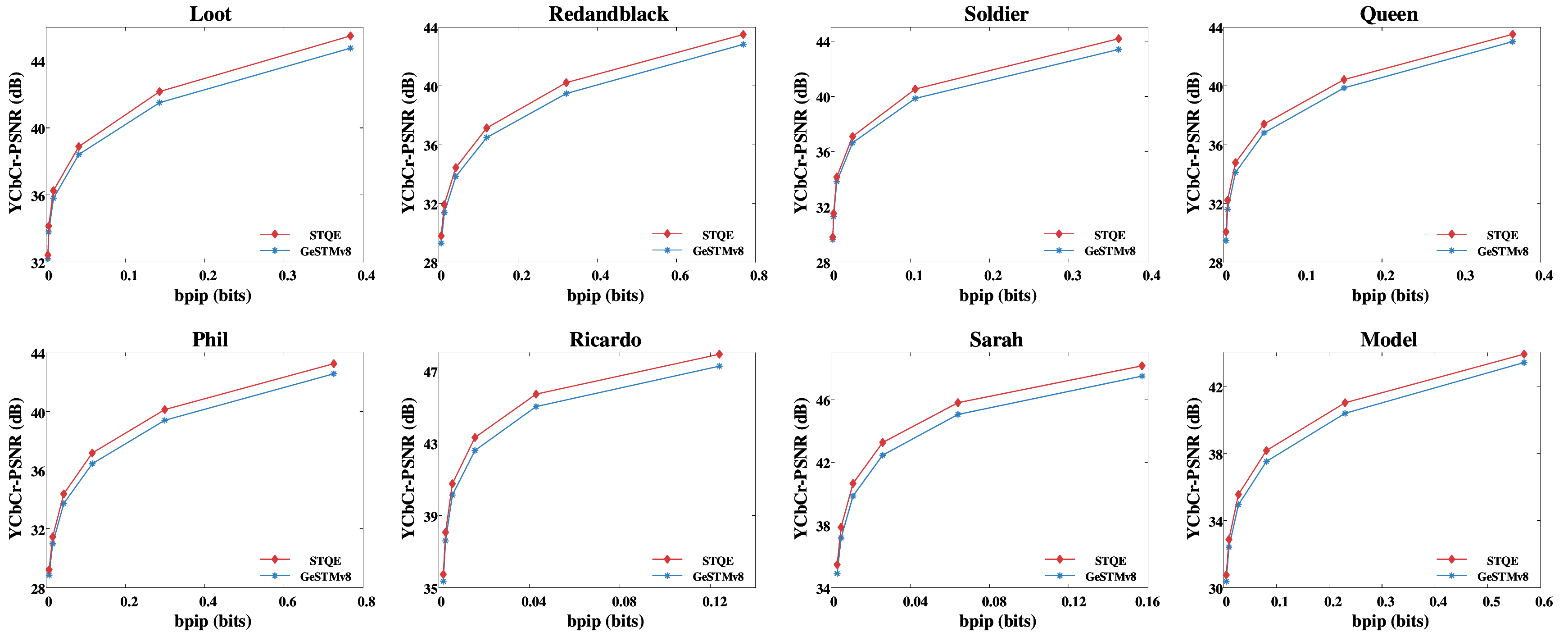}
\caption{Rate-PSNR curves before and after integrating STQE into GeSTMv8.}
\label{FIG9}
\end{figure*}

\subsection{Robustness Analysis}

\begin{table}
\centering
\caption{\(\Delta\)PSNR (dB) AND BD-rate (\%) AFTER INTEGRATING STQE INTO GESTMV8}
\label{tab:my_label1}  
\resizebox{8.55cm}{!}{
\begin{tabular}{lcccccccc}
\toprule
\multirow{2}{*}{\textbf{Sequence}} & \multicolumn{4}{c}{\textbf{\(\Delta\)PSNR (dB)}} & \multicolumn{4}{c}{\textbf{BD-rate (\%)}} \\ 
 & \textbf{Y} & \textbf{Cb} & \textbf{Cr} & \textbf{YCbCr} & \textbf{Y} & \textbf{Cb} & \textbf{Cr} & \textbf{YCbCr} \\ \hline
        Loot & 0.427 & 0.627 & 0.660 & 0.481 & -15.3 & -29.4 & -31.1 & -19.0 \\ 
        Redandblack & 0.638 & 0.474 & 0.660 & 0.620 & -20.2 & -25.9 & -17.8 & -20.6 \\ 
        Soldier & 0.411 & 0.516 & 0.555 & 0.442 & -13.6 & -29.2 & -29.6 & -17.6 \\ 
        Queen & 0.553 & 0.726 & 0.690 & 0.592 & -17.4 & -30.3 & -27.2 & -20.2 \\ 
        Dancer & 0.723 & 0.353 & 0.764 & 0.682 & -21.4 & -26.0 & -34.2 & -23.5 \\ 
        Model & 0.560 & 0.287 & 0.660 & 0.538 & -17.4 & -21.7 & -29.6 & -19.5 \\ 
        Phil & 0.709 & 0.319 & 0.266 & 0.605 & -19.5 & -19.4 & -13.6 & -18.7 \\ 
        Ricardo & 0.625 & 0.536 & 0.449 & 0.592 & -18.4 & -28.0 & -23.4 & -20.2 \\ 
        Sarah & 0.771 & 0.500 & 0.566 & 0.711 & -20.6 & -23.3 & -24.1 & -21.4 \\ \hline
        \textbf{Average} & \textbf{0.602} & \textbf{0.482} & \textbf{0.586} & \textbf{0.585} & \textbf{-18.2} & \textbf{-25.9} & \textbf{-25.6} & \textbf{-20.1} \\  \bottomrule
\end{tabular}}
\end{table}
\begin{table}
\centering
\caption{\(\Delta\)PSNR (dB) OF Y COMPONENT BY INTEGRATING STQE INTO GESTMV8}
\label{tab:my_label3}  
\resizebox{8cm}{!}{
\begin{tabular}{lcccccc}
\toprule
\multirow{2}{*}{\textbf{Sequence}} & \multicolumn{6}{c}{\textbf{\(\Delta\)PSNR (dB)}} \\ 
 & \textbf{R01} & \textbf{R02} & \textbf{R03} & \textbf{R04} & \textbf{R05} & \textbf{R06}  \\ \hline
        Loot & 0.202 & 0.270 & 0.362 & 0.392 & 0.610 & 0.728 \\ 
        Redandblack & 0.523 & 0.617 & 0.627 & 0.639 & 0.737 & 0.684 \\ 
        Soldier & 0.170 & 0.217 & 0.275 & 0.417 & 0.654 & 0.731 \\ 
        Queen & 0.613 & 0.578 & 0.610 & 0.518 & 0.517 & 0.481 \\ 
        Dancer & 0.615 & 0.782 & 0.903 & 0.823 & 0.680 & 0.537 \\ 
        Model & 0.414 & 0.488 & 0.643 & 0.678 & 0.643 & 0.495 \\ 
        Phil & 0.405 & 0.556 & 0.787 & 0.835 & 0.857 & 0.814 \\ 
        Ricardo & 0.401 & 0.499 & 0.669 & 0.767 & 0.729 & 0.685 \\ 
        Sarah & 0.611 & 0.707 & 0.862 & 0.873 & 0.832 & 0.739 \\ \hline
        \textbf{Average} & \textbf{0.439} & \textbf{0.524} & \textbf{0.638} & \textbf{0.660} & \textbf{0.695} & \textbf{0.655} \\ \bottomrule
\end{tabular}}
\end{table}
To further demonstrate the effectiveness of STQE, we integrate the above trained STQE models directly into a new in-developing 3D point cloud compression standard, i.e., Solid G-PCC, whose test platform is named as GeSTMv8 \cite{refb28}. All test sequences were compressed using GeSTMv8 with inter-frame prediction and octree-RAHT configuration. The average PSNRs and BD-rates are shown in Table III. We can see that STQE achieved average \(\Delta\)PSNR of 0.602 dB, 0.482 dB, 0.586 dB, and 0.585 dB for the Y, Cb, Cr components, and the combined YCbCr, respectively, corresponding to -18.2\%, -25.9\%, -25.6\%, and -20.1\% BD-rates, respectively. Table IV presents the \(\Delta\)PSNR of the Y-component at all six bitrates achieved by STQE. The gains were significant at medium and high bitrates, which is consistent with the results in Table II. Fig. 8 compares the rate-PSNR curves before and after integrating STQE into GeSTMv8. The results show that STQE also improved the coding efficiency of the Solid G-PCC encoder.
\begin{table*}
\centering
\caption{\(\Delta\)PSNR (dB) AND BD-rate (\%) COMPARISON BETWEEN GQE-NET \cite{refa2} AND STQE}
\label{tab:my_label8}  
\resizebox{15cm}{!}{
\begin{tabular}{lccccc|ccccc}
\toprule
\multirow{3}{*}{\textbf{Sequence}} & \multicolumn{5}{c}{\textbf{GQE-Net}} & \multicolumn{5}{c}{\textbf{STQE}}\\
& \multicolumn{4}{c}{\textbf{\(\Delta\)PSNR (dB)}} & {\textbf{BD-rate (\%)}} & \multicolumn{4}{c}{\textbf{\(\Delta\)PSNR (dB)}} & {\textbf{BD-rate (\%)}} \\ 
 & \textbf{Luma} & \textbf{Cb} & \textbf{Cr} & \textbf{YCbCr} & \textbf{Luma} &\textbf{Luma} & \textbf{Cb} & \textbf{Cr} & \textbf{YCbCr}& \textbf{Luma} \\ \hline
        Loot & 0.196 & 0.489 & 0.481 & 0.268 & -8.7 & 0.600 & 0.842 & 0.994 & 0.679 & -21.0 \\ 
        Redandblack & 0.239 & 0.273 & 0.294 & 0.250 & -8.3 & 0.785 & 0.665 & 0.857 & 0.779 & -23.3 \\ 
        Soldier & 0.278 & 0.355 & 0.432 & 0.307 & -16.0 & 0.811 & 0.550 & 0.607 & 0.753 & -26.8 \\ 
        Queen & 0.135 & 0.359 & 0.317 & 0.185 & -8.1 & 0.758 & 0.884 & 0.847 & 0.785 & -23.2 \\ 
        Dancer & 0.221 & 0.117 & 0.374 & 0.227 & -8.5 & 0.791 & 0.402 & 0.843 & 0.749 & -23.0 \\ 
        Model & 0.213 & 0.159 & 0.423 & 0.232 & -6.4 & 0.779 & 0.397 & 0.865 & 0.742 & -23.5 \\ 
        Phil & 0.202 & 0.134 & 0.156 & 0.188 & -3.7 & 0.883 & 0.357 & 0.496 & 0.769 & -23.4 \\ 
        Ricardo & 0.225 & 0.296 & 0.336 & 0.248 & -11.5 & 1.010 & 0.757 & 0.738 & 0.945 & -31.0 \\ 
        Sarah & 0.261 & 0.296 & 0.286 & 0.268 & -14.0 & 1.104 & 0.835 & 0.807 & 1.033 & -30.2 \\ \hline
        \textbf{Average} & 0.219 & 0.275 & 0.344 & 0.242 & -9.5 & \textbf{0.836} & \textbf{0.632} & \textbf{0.784} & \textbf{0.804} & \textbf{-24.9} \\ \bottomrule
\end{tabular}}
\vspace{4pt}
\end{table*}
\begin{figure*}
\centering
\includegraphics[width=6.1in]{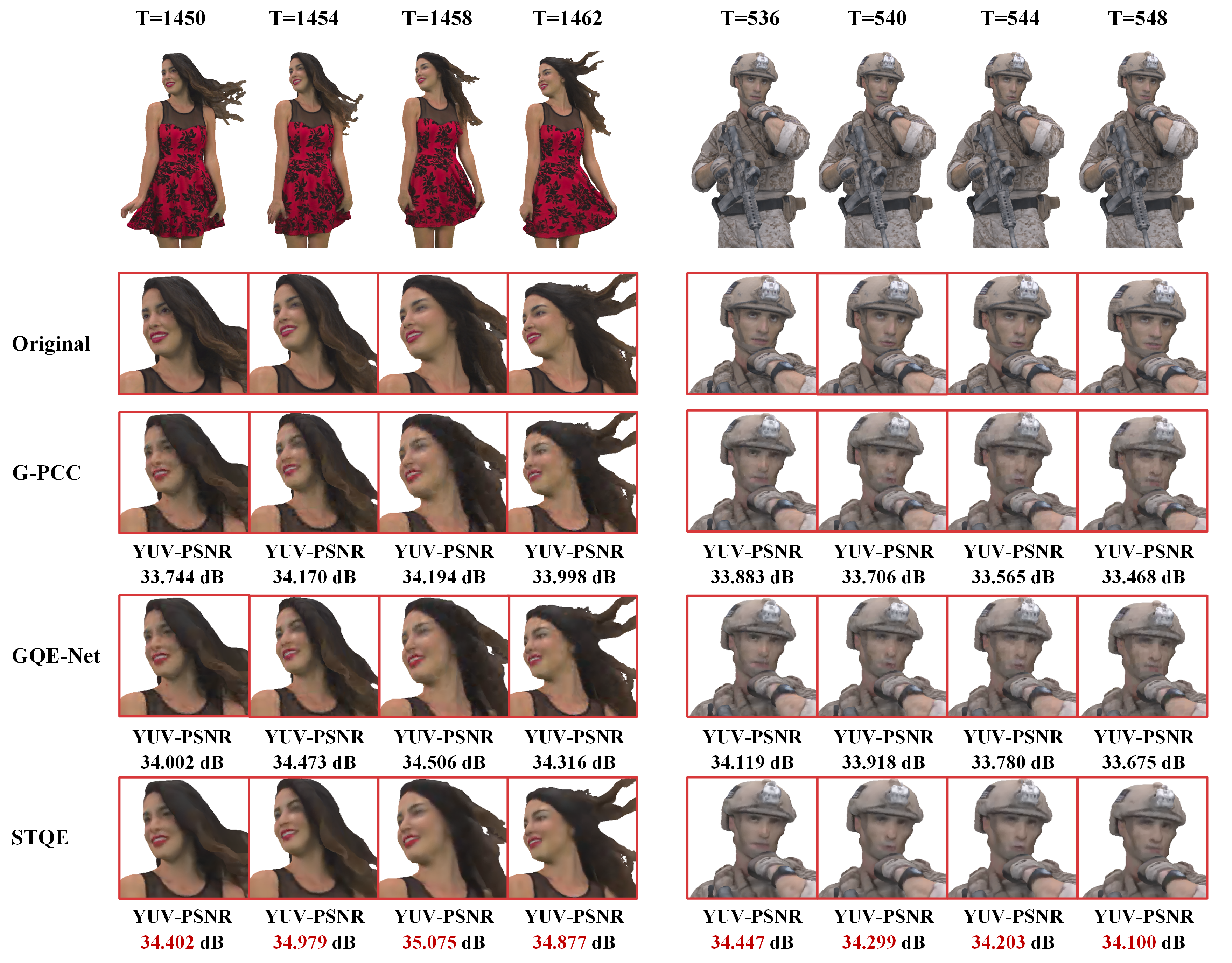}
\caption{Subjective quality comparison for the (from top to bottom) original point clouds, point clouds compressed and reconstructed by G-PCC, point clouds enhanced by GQE-Net, and point clouds enhanced by STQE, where T denotes the index of frame.}
\label{FIG8}
\end{figure*}

\subsection{Comparison with the State-of-the-Art}

To comprehensively evaluate the effectiveness of the proposed method, we compared it with GQE-Net \cite{refa2}, the current state-of-the-art learning-based point cloud quality enhancement method. We tested the first sixteen frames of each sequence. The average PSNRs and BD-rates of all tested sequences are given in Table V. The results show that the proposed method outperformed GQE-Net in terms of PSNR and coding efficiency, which is mainly attributed to the fact that GQE-Net failed to exploit the inter-frame correlation.

Taking sequences \textit{redandblack} and \textit{soldier} as an example, Fig. 9 compares the original point clouds in the first row, the point clouds compressed and reconstructed by G-PCC in the second row, the point clouds enhanced by GQE-Net in the third row, and the point clouds enhanced by STQE in the fourth row. Applying STQE significantly enhanced subjective quality, notably improving texture clarity and color transitions.

\begin{figure}
\centering
\includegraphics[width=3.4in]{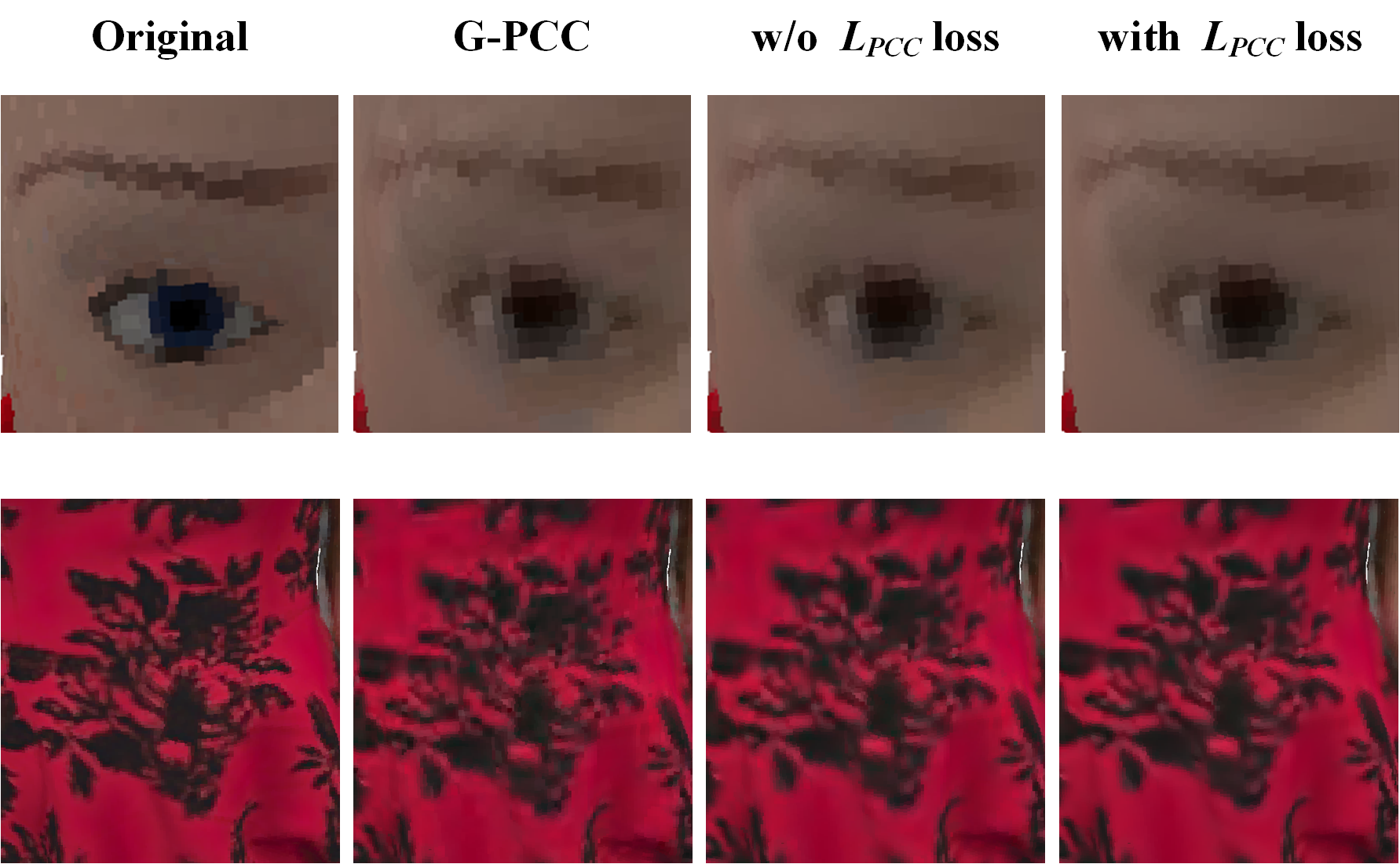}
\caption{Subjective quality comparison with and without \( L_{PCC}\) loss.}
\label{FIG10}
\end{figure}

\subsection{Ablation Study}

To verify the effectiveness of the proposed modules in STQE, we compared the performance of STQE with the following configurations:

\noindent (i) STQE \textbf{w/o RMC}, i.e., the RMC module was removed from STQE.

\noindent (ii) STQE \textbf{w/o CTA}, i.e., the CTA module was removed from STQE.

\noindent (iii) STQE \textbf{w/o GNFA}, i.e., we used an equivalent number of MLPs to replace GNFA.

\noindent (iv) STQE \textbf{w/o BID}, i.e., instead of using the two frames before and after the current frame as the reference frame, only the backward frame was used as the reference frame.

\noindent (v) STQE \textbf{w/o} \(\bm{L}_{{PCC}}\), i.e., \(\bm{L}_{{PCC}}\) was removed from the loss function during training.
\begin{table}
\centering
\caption{EFFECT OF RMC, CTA, BID, GNFA MODULES AND the \(\bm{L}_{PCC}\) LOSS IN TERMS OF \(\Delta\)PSNR}
\label{tab:my_label9} 
\resizebox{8.4cm}{!}{
\begin{tabular}{lcccccc}
\toprule
\textbf{\shortstack{Sequence \\ [-0.25ex]~}} & 
\textbf{\shortstack{w/o \\ RMC}} & 
\textbf{\shortstack{w/o \\ CTA}} & 
\textbf{\shortstack{w/o \\ GNFA}} & 
\textbf{\shortstack{w/o \\ BID}} & 
\textbf{\shortstack{w/o \\ \(\bm{L}_{PCC}\)}} & 
\textbf{\shortstack{STQE \\ [-0.25ex]~}} \\ \hline
        Loot & 0.540 & 0.575 & 0.438 & 0.515 & 0.616 & 0.649 \\ 
        Redandblack & 0.710 & 0.707 & 0.562 & 0.709  & 0.796 & 0.830 \\ 
        Soldier & 1.059 & 1.032 & 0.879 & 0.999  & 1.112 & 1.202 \\ 
        Queen & 0.656 & 0.693 & 0.552 & 0.651 & 0.746 & 0.796 \\ 
        Dancer & 0.841 & 0.807 & 0.655 & 0.878  & 0.913 & 0.955 \\ 
        Model & 0.782  & 0.783 & 0.659 & 0.807 & 0.874 & 0.881 \\ 
        Phil & 0.855  & 0.923 & 0.757 & 0.889 & 0.998 & 1.033 \\ 
        Ricardo & 0.820  & 0.966  & 0.814  & 0.966  & 1.073 & 1.170 \\ 
        Sarah & 0.818  & 0.975  & 0.816  & 0.984  & 1.087 & 1.146 \\ \hline
        \textbf{Average} & 0.787 & 0.829 & 0.681 & 0.822  & 0.913 & \textbf{0.962 }\\  \bottomrule
\end{tabular}}
\end{table}

Table VI shows the results. All the modules (RMC, CTA, BID, and GNFA), as well as the \(L_{PCC}\) loss, improved the overall performance of STQE. The RMC module supported effective temporal feature extraction through accurate inter-frame motion compensation. Without the RMC module, the average PSNR gain decreased by 0.175 dB. The CTA module improved performance by focusing on reference regions with higher relevance to the current frame and selecting them effectively. This approach increased the average PSNR gain by 0.133 dB. On the other hand, the GNFA module led to an average PSNR gain of 0.281 dB. This improvement occurred because the module extracted spatial features efficiently based on the distribution pattern of the point cloud. Introducing bidirectional reference frames increased the \(\Delta\)PSNR from 0.822 dB to 0.962 dB. Since the \(\Delta\)PSNR gain contributed by the \(L_{PCC}\) loss was relatively small, we compared the subjective quality with and without it in Fig. 10. The eyebrow and eye details of \textit{Queen}, as well as the skirt pattern of \textit{Redandblack} show that STQE with \(L_{PCC}\) retained high-frequency details. In contrast, STQE without \(L_{PCC}\) led to over-smoothing artifacts. 

\subsection{Computational Complexity Analysis}

Table VII compares the computational complexity of STQE and GQE-Net in terms of parameters, floating-point operations (FLOPs), and average processing time for all test sequences. The results show that GQE-Net has 0.59M parameters, 34.85G FLOPs, and an average processing time of 95.71s, whereas STQE has only 0.36M parameters, 20.07G FLOPs, and processing time of 25.19s.
\begin{table}
\centering
\caption{COMPUTATIONAL COMPLEXITY COMPARISON BETWEEN GQE-NET \cite{refa2} AND STQE}
\label{tab:my_label7}  
\resizebox{8.4cm}{!}{
\begin{tabular}{lccc}
\toprule
        \textbf{Method} & \textbf{Processing time (s)} & \textbf{FLOPs (G)} & \textbf{Parameters (M)} \\ \hline
        GQE-Net & 95.71 & 34.85 & 0.59 \\ 
        STQE & 25.19 & 20.07 & 0.36 \\ \bottomrule
\end{tabular}}
\end{table}

\section{Conclusion}
We proposed STQE, a spatial-temporal attribute quality enhancement method for G-PCC compressed dynamic point clouds, consisting of three novel modules: RMC, CTA and GNFA. The RMC module accurately aligns inter-frame geometry coordinates, addressing challenges caused by varying number of points and inter-frame motion. The CTA module dynamically focuses on reference frames that are more relevant to the current frame. The GNFA module uses the statistical distribution of distance and color attributes in the point cloud to adaptively assign larger weights to the features in the neighborhood that have higher correlation with the current point. Moreover, we introduced a Pearson correlation coefficient-based loss as supplementary supervision to effectively restore texture details. Experimental results demonstrate that STQE improves the quality of the compressed dynamic point clouds significantly. In the future, we aim to address the quality fluctuation problem among frames and further reduce the complexity.

% \section{References Section}
% You can use a bibliography generated by BibTeX as a .bbl file.
%  BibTeX documentation can be easily obtained at:
%  http://mirror.ctan.org/biblio/bibtex/contrib/doc/
%  The IEEEtran BibTeX style support page is:
%  http://www.michaelshell.org/tex/ieeetran/bibtex/
 
%  % argument is your BibTeX string definitions and bibliography database(s)
% %\bibliography{IEEEabrv,../bib/paper}
% You can manually copy in the resultant .bbl file and set second argument of $\backslash${\tt{begin}} to the number of references
%  (used to reserve space for the reference number labels box).

\begin{IEEEbiography}[{\includegraphics[width=1in,height=1.25in,clip,keepaspectratio]{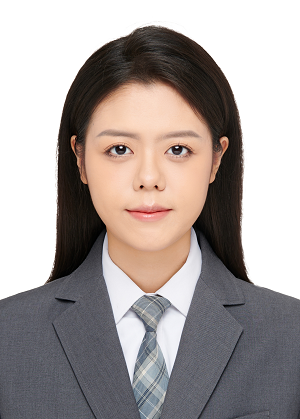}}]
 {Tian Guo}
 received the B.E. degree from the School of Information and Control Engineering, China University of Mining and Technology, Jiangsu, China, in 2021. She is currently pursuing the Ph.D. degree with the School of Control Science and Engineering, Shandong University, Shandong, China. Her research interests include point cloud compression and processing.
 \end{IEEEbiography}

 \begin{IEEEbiography}[{\includegraphics[width=1in,height=1.25in,clip,keepaspectratio]{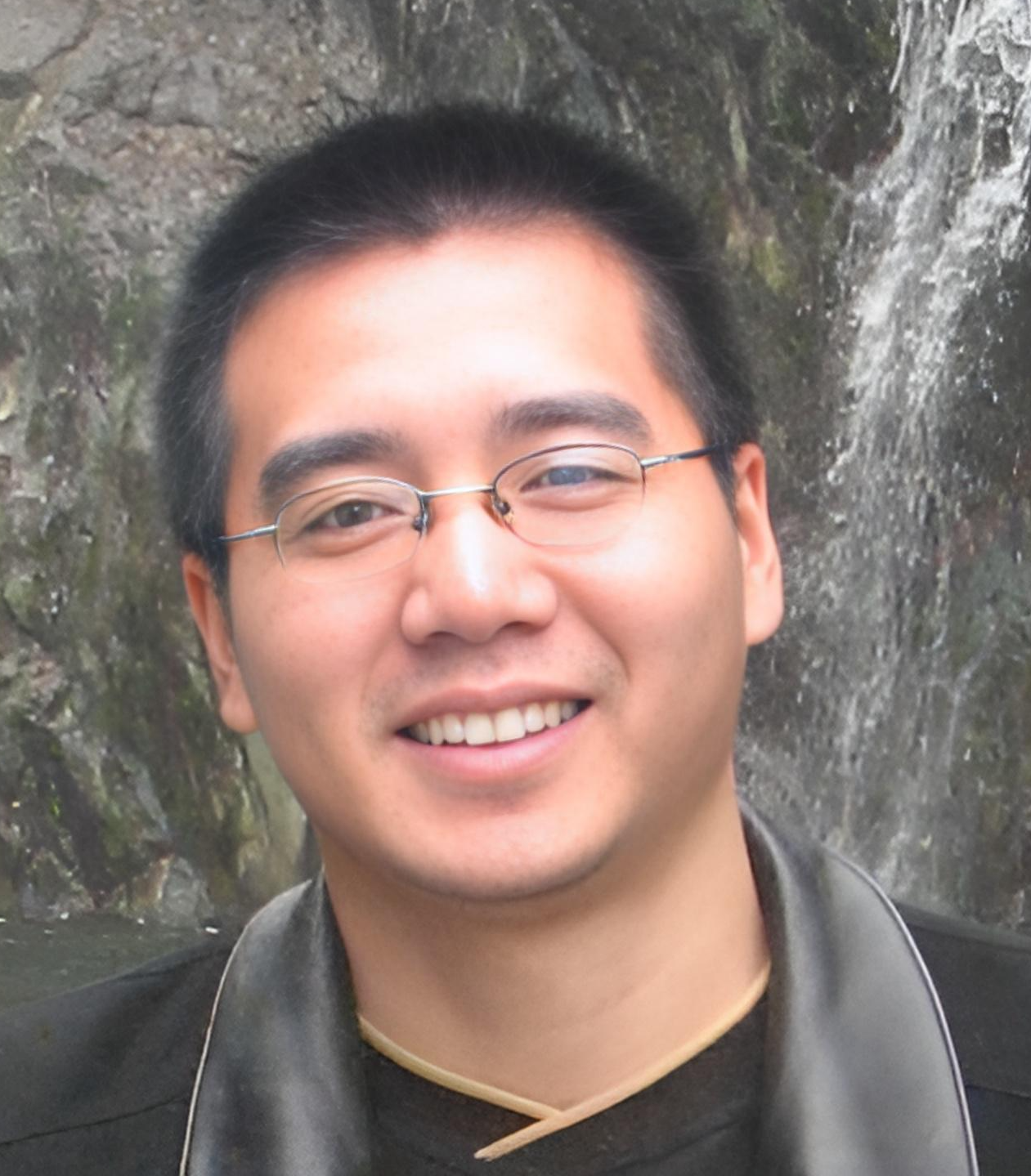}}]
 {Hui Yuan}
(Senior Member, IEEE) received the B.E. and Ph.D. degrees in telecommunication engineering from Xidian University, Xi’an, China, in 2006 and 2011, respectively. In April 2011, he joined Shandong University, Jinan, China, as a Lecturer (April 2011–December 2014), an Associate Professor (January 2015–August 2016), and a Professor (September 2016). From January 2013 to December 2014 and from November 2017 to February 2018, he was a Postdoctoral Fellow (Granted by the Hong Kong Scholar Project) and a Research Fellow, respectively, with the Department of Computer Science, City University of Hong Kong. From November 2020 to November 2021, he was a Marie Curie Fellow (Granted by the Marie Skłodowska-Curie Actions Individual Fellowship under Horizon2020 Europe) with the School of Engineering and Sustainable Development, De Montfort University, Leicester, U.K. From October 2021 to November 2021, he was also a Visiting Researcher (secondment of the Marie Skłodowska-Curie Individual Fellowships) with the Computer Vision and Graphics Group, Fraunhofer Heinrich-Hertz-Institut (HHI), Germany. His current research interests include 3D visual coding, processing, and communication. He is also serving as an Area Chair for IEEE ICME, an Associate Editor for \textit{IEEE Transactions on Image Processing}, \textit{IEEE Transactions on Consumer Electronics}, and \textit{IET Image Processing}.
 \end{IEEEbiography}

 \begin{IEEEbiography}[{\includegraphics[width=1in,height=1.25in,clip,keepaspectratio]{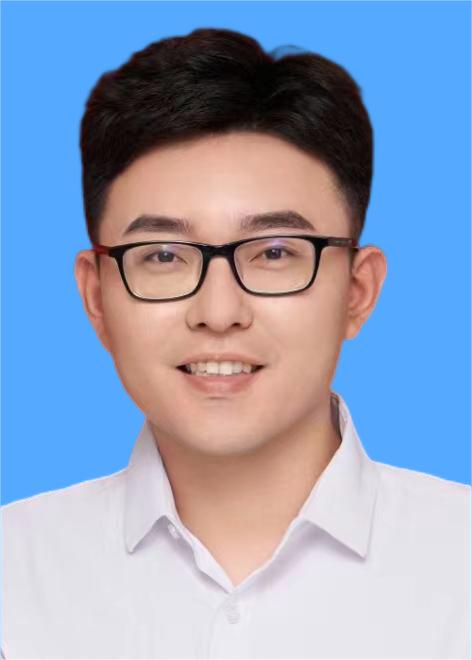}}]
 {Xiaolong Mao}
received the M.E. degree from the School of Integrated Circuits, Shandong University, Shandong, China, in 2021.He is currently pursuing a Ph.D. degree at Shandong University. His research interests include point clouds compression and processing.
 \end{IEEEbiography}

 \begin{IEEEbiography}[{\includegraphics[width=1in,height=1.25in,clip,keepaspectratio]{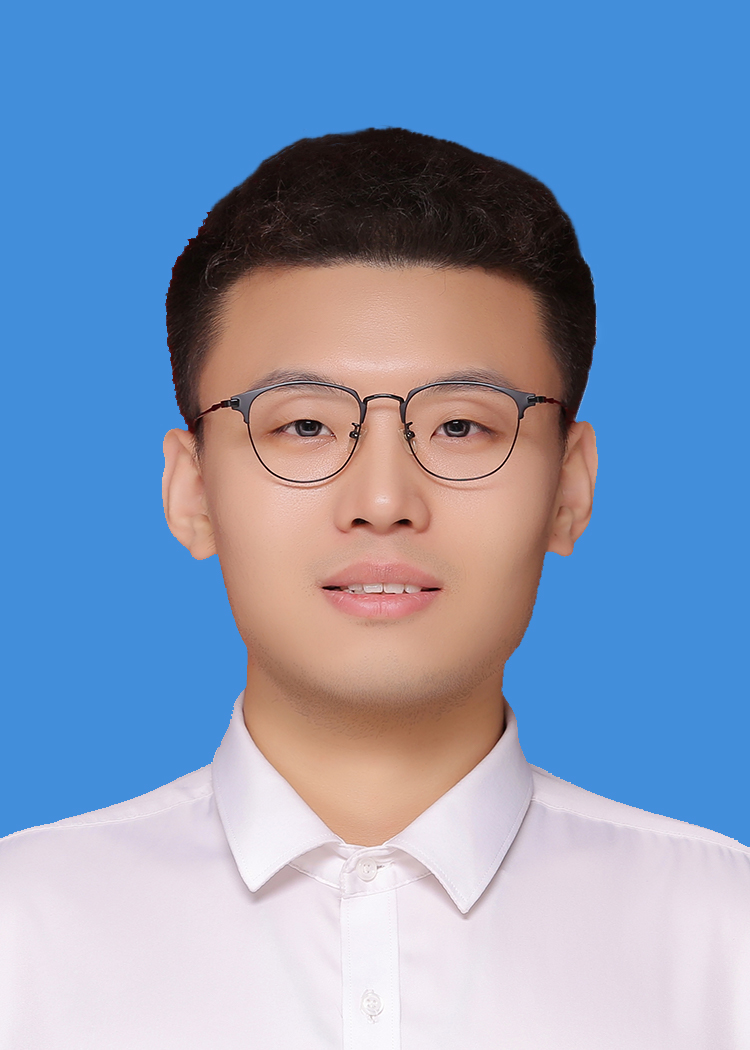}}]
 {Shiqi Jiang}
 is currently pursuing the Ph.D. degree in artificial intelligence with the School of Software, Shandong University, Jinan, China. His research interest interests computer vision, image and video coding/processing, and deep learning.
 \end{IEEEbiography}

 \begin{IEEEbiography}[{\includegraphics[width=1in,height=1.25in,clip,keepaspectratio]{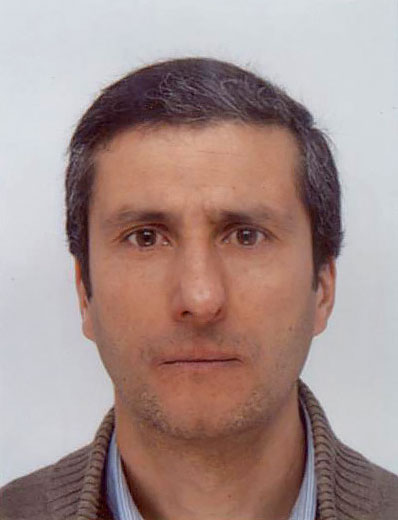}}]
 {Raouf Hamzaoui}
(Senior Member, IEEE) received the M.Sc. degree in mathematics from the University of Montreal, Canada, in 1993, and the Dr.rer.nat. degree from the University of Freiburg, Germany, in 1997, and the Habilitation degree in computer science from the University of Konstanz, Germany, in 2004. He was an Assistant Professor with the Department of Computer Science, University of Leipzig, Germany, and the Department of Computer and Information Science, University of Konstanz. In September 2006, he joined De Montfort University, where he is currently a Professor in media technology. He was a member of the Editorial Board of the IEEE TRANSACTIONS ON MULTIMEDIA and IEEE TRANSACTIONS ON CIRCUITS AND SYSTEMS FOR VIDEO TECHNOLOGY. He has published more than 120 research papers in books, journals, and conferences. His research has been funded by the EU, DFG, Royal Society, and industry and received best paper awards (ICME 2002, PV’07, CONTENT 2010, MESM’2012, and UIC-2019).
 \end{IEEEbiography}

 \begin{IEEEbiography}[{\includegraphics[width=1in,height=1.25in,clip,keepaspectratio]{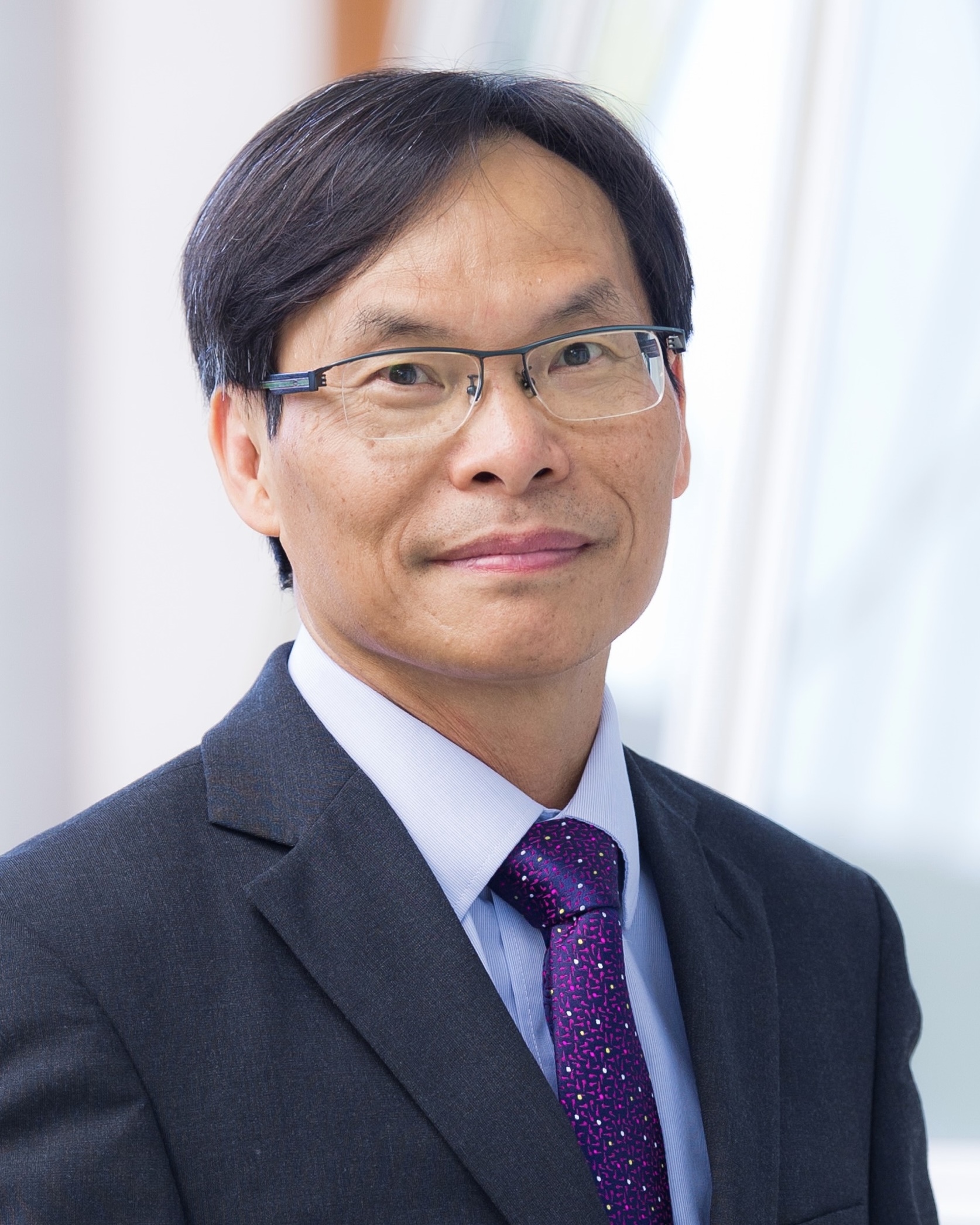}}]
 {Sam Kwong}
(Fellow, IEEE) is the Chair Professor of Computational Intelligence and concurrently serves as the Associate Vice-President (Strategic Research) at Lingnan University. He received the B.S. degree from the State University of New York at Buffalo in 1983, the M.S. degree in electrical engineering from the University of Waterloo, Canada, in 1985, and the Ph.D. degree from the University of Hagen, Germany, in 1996. From 1985 to 1987, he was a Diagnostic Engineer with Control Data Canada. He then joined Bell Northern Research Canada. Since 1990, he has been with City University of Hong Kong, where he served as a Lecturer in the Department of Electronic Engineering and later became a Chair Professor in the Department of Computer Science before moving to Lingnan University in 2023. His research interests include video/image coding, evolutionary algorithms, and artificial intelligence solutions. He is a Fellow of the IEEE, the Hong Kong Academy of Engineering Sciences (HKAES), and the National Academy of Inventors (NAI), USA. Dr. Kwong was honored as an IEEE Fellow in 2014 for contributions to optimization techniques in cybernetics and video coding and was named a Clarivate Highly Cited Researcher in 2022. He currently serves as an Associate Editor for the IEEE Transactions on Industrial Electronics and the IEEE Transactions on Industrial Informatics, among other prestigious IEEE journals. He has authored over 350 journal papers and 160 conference papers, achieving an h-index of 93 (Google Scholar). He served as President of the IEEE Systems, Man, and Cybernetics Society (SMCS) from 2021 to 2023.
 \end{IEEEbiography}
% \newpage

 

% \end{IEEEbiography}



\vfill

\end{document}